\documentclass[conference]{IEEEtran}
\IEEEoverridecommandlockouts
\usepackage{amssymb}
\usepackage{hyperref}
\usepackage{amsmath,bm}
\usepackage{algorithm}
\usepackage{algpseudocode}
\usepackage{cite}
\usepackage{tabularx}
\usepackage{tabulary}
\usepackage{mathrsfs}
\usepackage{graphicx}
\usepackage{epstopdf}
\usepackage{multirow}
\usepackage{makecell}
\usepackage{booktabs}
\def\BibTeX{{\rm B\kern-.05em{\sc i\kern-.025em b}\kern-.08em
    T\kern-.1667em\lower.7ex\hbox{E}\kern-.125emX}}
\begin{document}

\title{Adaptive sparseness for correntropy-based\\robust regression via automatic\\relevance determination\\
\thanks{This work was supported in part by the Japan Society for the Promotion of Science (JSPS) KAKENHI under Grant 19H05728, in part by the Japan Science and Technology Agency (JST) Support for the Pioneering Research Initiated by Next Generation (SPRING) under Grant JPMJSP2106, and in part by the National Natural Science Foundation of China under Grant U21A20485 and Grant 61976175. \emph{(*Corresponding author: Yuanhao Li.)}}
\thanks{Yuanhao Li, Natsue Yoshimura, and Yasuharu Koike are with the Institute of Innovative Research, Tokyo Institute of Technology, Yokohama 226-8503, Japan. (correspondence e-mail: li.y.ay@m.titech.ac.jp)}
\thanks{Badong Chen is with the Institute of Artificial Intelligence and Robotics, Xi'an Jiaotong University, Xi'an 710049, China.}
\thanks{Okito Yamashita is with the Department of Computational Brain Imaging, ATR Neural Information Analysis Laboratories, Kyoto 619-0288, Japan.}}
\author{\IEEEauthorblockN{Yuanhao Li*, Badong Chen, Okito Yamashita, Natsue Yoshimura, and Yasuharu Koike}}
\maketitle
\begin{abstract}
Sparseness and robustness are two important properties for many machine learning scenarios. In the present study, regarding the \emph{maximum correntropy criterion} (MCC) based robust regression algorithm, we investigate to integrate the MCC method with the \emph{automatic relevance determination} (ARD) technique in a Bayesian framework, so that MCC-based robust regression could be implemented with `\emph{adaptive sparseness}'. To be specific, we use an inherent noise assumption from the MCC to derive an explicit likelihood function, and realize the maximum a posteriori (MAP) estimation with the ARD prior by variational Bayesian inference. Compared to the existing robust and sparse $L_1$-regularized MCC regression, the proposed MCC-ARD regression can eradicate the troublesome tuning for the regularization hyper-parameter which controls the regularization strength. Further, MCC-ARD achieves superior prediction performance and feature selection capability than $L_1$-regularized MCC, as demonstrated by a noisy and high-dimensional simulation study.
\end{abstract}


\section{Introduction}
\label{sec:introduction}
\noindent Regression aims at a prediction model for continuous variables from the input of covariate variables or some derived features, which is also closely related to system identification, adaptive filtering, and so on. Consider the following canonical linear-in- parameter (LIP) model with additive noise
\begin{equation}
	\label{equ:lip}
	t=\varPhi(\mathbf{x})\mathbf{w}+\epsilon
\end{equation}
where $t$ denotes the model output, $\varPhi(\mathbf{x})$ is a mapping of input $\mathbf{x}$, $\mathbf{w}$ is the model parameter, while $\epsilon$ denotes the noise term. If we exclude the utilization of the mapping function $\varPhi(\cdot)$, LIP model degenerates to the linear regression model
\begin{equation}
	\label{equ:lr}
	t=\mathbf{x}\mathbf{w}+\epsilon
\end{equation}
in which one can suppose that $\mathbf{x}=(x_1,x_2,\cdots,x_D)\in \mathbb{R}^{1\times D}$ is the $D$-dimensional covariate while $\mathbf{w}=(w_1,w_2,\cdots,w_D)^T$ $\in \mathbb{R}^{D\times 1}$ is the model parameter. $T$ denotes the transpose for a vector or matrix. The most common method for learning $\mathbf{w}$ is to minimize the expectation of the quadratic error $e\triangleq t-\mathbf{x}\mathbf{w}$ which refers to the least square (LS) criterion
\begin{equation}
	\label{equ:lsc}
	\mathbf{w}=arg\min _{\mathbf{w}} \left< e^2 \right>=arg\min _{\mathbf{w}} \left< (t-\mathbf{x}\mathbf{w})^2 \right>
\end{equation}
where $\left< \cdot\right>$ denotes the mathematical expectation. However, the traditional least square method is only effective for well-posed questions. When $D>N$, (\ref{equ:lsc}) will result in poor generalization performance. A useful solution is to select a subset of features while pruning those irrelevant features, which is called sparse learning. In the learned model parameter $\mathbf{w}$, many components will be zero so that the corresponding features are pruned. The idealized sparse model is to minimize the $L_0$-regularized cost function
\begin{equation}
	\label{equ:lsc_l0}
	\mathbf{w}=arg\min _{\mathbf{w}} \left< e^2 \right>+\lambda\|\mathbf{w}\|_0
\end{equation}
where $\lambda$ is hyper-parameter tuning the regularization strength, while $\|\mathbf{w}\|_0$ is the $L_0$-norm of $\mathbf{w}$ denoting the number of non-zero components in $\mathbf{w}$. Since solving (\ref{equ:lsc_l0}) is NP-hard, $L_0$-norm is usually replaced with its tightest \emph{convex} relaxation $L_1$-norm \cite{wipf2007new} which leads to the LASSO algorithm \cite{tibshirani1996regression}
\begin{equation}
	\label{equ:lsc_l1}
	\mathbf{w}=arg\min _{\mathbf{w}} \left< e^2 \right>+\lambda\|\mathbf{w}\|_1
\end{equation}
which has been well studied and discussed for sparse learning \cite{figueiredo2003adaptive,krishnapuram2004bayesian,krishnapuram2005sparse,schmidt2007fast}. However, the hyper-parameter $\lambda$ is usually a nuisance which would require manual tuning or time-consuming cross-validation.

An alternative way to solve a sparse model is the automatic relevance determination (ARD) technique \cite{mackay1992practical}, which has been receiving growing attention with the proposal of the relevance vector machine (RVM) \cite{tipping1999relevance,tipping2001sparse,bishop2000variational}, a Bayesian treating of support vector machine (SVM). ARD supposes a prior distribution for $\mathbf{w}$ with a hierarchical form, and infers the posterior distribution for $\mathbf{w}$, combining with the likelihood function, in the Bayesian framework. ARD has proved as a tighter approximation of $L_0$-norm than $L_1$-norm, thus providing superior sparse capability, although it is \emph{non-convex} in the regularization form \cite{wipf2007new}. More importantly, ARD could infer all the unknown variables, while excluding the regularization hyper-parameter $\lambda$, thus realizing `\emph{adaptive sparseness}'.   

On the other hand, the least-square criterion implicitly uses a Gaussian assumption on the noise $\epsilon$, which need not be the truth in practice. In particular, least-square methods can suffer serious degeneration in the presence of outliers. The \emph{maximum correntropy criterion} (MCC) is highly efficient for noisy data analysis \cite{liu2007correntropy,feng2015learning,chen2016generalized,ma2018bias}, which has been also used for robust sparse learning integrating with $L_1$-regularization \cite{he2010maximum,he2011regularized,he2013half} or other regularization terms \cite{ma2015maximum,lu2020robust}. Yet, as mentioned before, they need careful tuning on the regularization hyper-parameters. In this work, we desire to introduce the Bayesian ARD technique to the MCC-based robust regression for `\emph{adaptive sparseness}', which remains a vacancy in the literature.

The remainder of this paper is organized as follows. Section \ref{sec:lsr_ard} reviews the ARD-based sparse regression algorithm with the Gaussian assumption for the noise term $\epsilon$. In Section \ref{sec:mcc}, we give a brief introduction about MCC and show the assumption on the noise distribution when MCC is used as the regression objective function. In Section \ref{sec:mcc-ard}, we propose to employ MCC as the likelihood function with ARD technique in the Bayesian framework for robust sparse regression. In Section \ref{sec:exp}, we show some experimental results to demonstrate the superiority of the proposed method. In Section \ref{sec:disc}, we provide some discussions. Finally, Section \ref{sec:con} concludes this paper.

\section{ARD-Based Sparse Regression}
\label{sec:lsr_ard}
Supposing the zero-mean Gaussian distribution for the noise term with the variance being $\sigma^2$, we can obtain the probability density function (PDF) for $t$ by $p(t|\mathbf{x})=\mathcal{N}(t|\mathbf{xw},\sigma^2)$, which is a Gaussian distribution over $t$ with mean $\mathbf{xw}$ and variance $\sigma^2$. With an input-target dataset $\{\mathbf{x}_n,t_n\}_{n=1}^N$ and assuming the independence of $t_n$, we could write the likelihood function
\begin{equation}
	\label{equ:likelihood}
	p(\mathbf{t}|\mathbf{w},\sigma^2)=(2\pi\sigma^2)^{-N/2}\exp\{-\frac{1}{2\sigma^2}\|\mathbf{t}-\mathbf{X}\mathbf{w}\|^2\}
\end{equation}
in which $\mathbf{t}=(t_1,t_2,\cdots,t_N)^T\in \mathbb{R}^{N\times 1}$, $\mathbf{X}\in \mathbb{R}^{N\times D}$ denotes the collection for $\mathbf{x}_n$, each row of which represents a sample. For simplicity, the dependence upon the covariate matrix $\mathbf{X}$ is omitted in ($\ref{equ:likelihood}$) and also subsequent expressions. The maximum likelihood estimation (MLE) of ($\ref{equ:likelihood}$) is equal to the least square criterion, which exhibits the following closed-form solution
\begin{equation}
	\label{equ:ls}
	\mathbf{w}=(\mathbf{X}^T\mathbf{X})^{-1}\mathbf{X}^T\mathbf{t}
\end{equation}
If $D>N$, the solution (\ref{equ:ls}) will be ill-posed. To select a subset of features for the regression task, one could employ the ARD technique that assigns the zero-mean and anisotropic Gaussian distribution for each model parameter with individual inverse variances $\mathbf{a}=(a_1,a_2,\cdots,a_D)$
\begin{equation}
	\label{equ:ard1}
	p(\mathbf{w}|\mathbf{a})=\prod_{d=1}^{D}p(w_d|a_d) = \prod_{d=1}^{D}\mathcal{N}(w_d|0,a_d^{-1})
\end{equation}
where $a_d$ (the inverse variance) is called relevance parameter, which controls the possible range for corresponding $w_d$. Each relevance parameter is then assigned with the non-informative Jeffreys hyper-prior (which is actually an \emph{improper} prior\footnote{Note that this prior is in fact an \emph{improper} prior since it is not normalizable (the integral is infinite).}\cite{gelman1995bayesian})
\begin{equation}
	\label{equ:ard2}
	p(\mathbf{a})=\prod_{d=1}^{D}p(a_d) =\prod_{d=1}^{D} a_d^{-1}
\end{equation}
The prior distribution for noise variance $\sigma^2$ is usually assumed to be non-informative as well
\begin{equation}
	\label{equ:ard3}
	p(\sigma^2) = (\sigma^2)^{-1}
\end{equation}
Having defined the likelihood and also the prior in (\ref{equ:likelihood})(\ref{equ:ard1})-(\ref{equ:ard3}), we can write analytically the posterior distribution over $\mathbf{w}$
\begin{equation}
	\begin{split}
		\label{equ:post_w}
		&p(\mathbf{w}|\mathbf{t},\mathbf{a},\sigma^2)=\frac{p(\mathbf{t}|\mathbf{w},\sigma^2)p(\mathbf{w}|\mathbf{a})}{p(\mathbf{t}|\mathbf{a},\sigma^2)}=\frac{p(\mathbf{t}|\mathbf{w},\sigma^2)p(\mathbf{w}|\mathbf{a})}{\int p(\mathbf{t}|\mathbf{w},\sigma^2)p(\mathbf{w}|\mathbf{a})d\mathbf{w}} \; \\
		&=(2\pi)^{-D/2}|\mathbf{\Sigma}|^{-1/2}\exp\{-\frac{1}{2}(\mathbf{w}-\boldsymbol{\mu})^T\mathbf{\Sigma}^{-1}(\mathbf{w}-\boldsymbol{\mu})\} \; \\
	\end{split}
\end{equation}
in which the covariance and mean for $\mathbf{w}$ are computed by
\begin{equation}
	\begin{split}
		\label{equ:post_w_2}
		\mathbf{\Sigma}&=(\sigma^{-2}\mathbf{X}^T\mathbf{X}+\mathbf{A})^{-1} \; \\
		\boldsymbol{\mu}&=\sigma^{-2}\mathbf{\Sigma}\mathbf{X}^T\mathbf{t} \; \\
	\end{split}
\end{equation}
with $\mathbf{A}=diag(a_1,a_2,\cdots,a_D)$. To obtain the whole posterior distribution
\begin{equation}
	\label{equ:post_whole}
	p(\mathbf{w},\mathbf{a},\sigma^2|\mathbf{t}) = p(\mathbf{w}|\mathbf{t},\mathbf{a},\sigma^2)p(\mathbf{a},\sigma^2|\mathbf{t})
\end{equation}
one notes that the hyper-parameter posterior distribution could be denoted by $p(\mathbf{a},\sigma^2|\mathbf{t})\propto p(\mathbf{t}|\mathbf{a},\sigma^2)p(\mathbf{a})p(\sigma^2)$. Utilizing the non-informative hyper-priors, we only need to optimize $\mathbf{a}$ and $\sigma^2$ so that the \emph{marginal likelihood} $p(\mathbf{t}|\mathbf{a},\sigma^2)$ is maximized
\begin{equation}
	\begin{split}
		\label{equ:marginal_likelihood}
		p(\mathbf{t}|\mathbf{a},\sigma^2)=&\int p(\mathbf{t}|\mathbf{w},\sigma^2)p(\mathbf{w}|\mathbf{a})d\mathbf{w} \; \\
		=&(2\pi)^{-D/2}|\sigma^2\mathbf{I}+\mathbf{X}\mathbf{A}^{-1}\mathbf{X}^T|^{-1/2}  \; \\
		&\times\exp\{-\frac{1}{2}\mathbf{t}^T(\sigma^2\mathbf{I}+\mathbf{X}\mathbf{A}^{-1}\mathbf{X}^T)^{-1}\mathbf{t}\} \; \\
	\end{split}
\end{equation}
To maximize (\ref{equ:marginal_likelihood}), setting the differentiation to zero yields the following update
\begin{equation}
	\label{equ:lsr_ard_a}
	a_d = \frac{\gamma_d}{\mu_d^2}
\end{equation}
in which $\mu_d$ is the $d$-th component of $\boldsymbol{\mu}$ and $\gamma_d$ is defined by $\gamma_d\triangleq1-a_d\Sigma_{dd}$ with $\Sigma_{dd}$ the $d$-th diagonal element of $\mathbf{\Sigma}$. $\sigma^2$ is updated by
\begin{equation}
	\label{equ:lsr_ard_b}
	\sigma^2 = \frac{\|\mathbf{t}-\mathbf{X}\boldsymbol{\mu}\|^2}{N-\sum_{d=1}^{D}\gamma_d}
\end{equation}
Updating (\ref{equ:post_w_2})(\ref{equ:lsr_ard_a})(\ref{equ:lsr_ard_b}) alternately, we will obtain the \emph{maximum a posteriori} (MAP) estimations for all the unknown variables. In particular, during the inference, those $a_d$ which correspond to irrelevant features will diverge to arbitrarily large numbers, so that the probability density of the corresponding $w_d$ focuses at the origin, thus pruning the irrelevant features and realizing sparse regression.

The above-described optimization involves maximization of the \emph{marginal likelihood} $p(\mathbf{t}|\mathbf{a},\sigma^2)$ (\ref{equ:marginal_likelihood}), which is known as the \emph{type-II maximum likelihood} \cite{gelman1995bayesian}. Moreover, the model can be optimized in other ways. For example, Expectation-Maximum (EM) could be employed by regarding the relevance parameter $\mathbf{a}$ as the hidden variables \cite{figueiredo2003adaptive}. One could also use the variational Bayesian (VB) method with surrogate function to approximate the posterior distribution for every random variable \cite{bishop2000variational}. Since the conventional ARD-based sparse regression is derived under the assumption of Gaussian noise (\ref{equ:likelihood}), it may suffer significant performance degeneration in a realistic non-Gaussian scenario, in particular in the presence of outliers \cite{liu2007correntropy,ma2018bias,chen2018common,xing2018robust}.

\section{Maximum Correntropy Criterion}
\label{sec:mcc}

\subsection{Maximum Correntropy Criterion}
\label{subsec:mcc}
Correntropy was originally developed as a generalized form of correlation function for stochastic processes, which has been further extended as a similarity measure between two arbitrary variables for machine learning and signal processing \cite{liu2007correntropy}. For two variables $A$ and $B$ with joint distribution $p_{A,B}(a,b)$, their correntropy similarity is defined by
\begin{equation}
	\label{equ:correntropy}
	\mathcal{V}(A,B)\triangleq\left< k(A,B)\right>=\int k(a,b) dp_{A,B}(a,b)
\end{equation}
where $k(\cdot,\cdot)$ is a shift-invariant \emph{Mercer} kernel which is usually implemented with the Gaussian kernel function
\begin{equation}
	\label{equ:gaussian_kernel}
	k_h(a,b)\triangleq \exp(-\frac{(a-b)^2}{2h})
\end{equation}
where $h>0$ denotes the kernel bandwidth, controlling all the robust property for correntropy. Given $N$ samples of variables $A$ and $B$, the empirical estimation of correntropy is computed by
\begin{equation}
	\begin{split}
		\label{equ:correntropy_est}
		\hat{\mathcal{V}}(A,B)&=\frac{1}{N}\sum_{n=1}^{N}{k_h(a_n,b_n)} \; \\
		&=\frac{1}{N}\sum_{n=1}^{N}{\exp(-\frac{(a_n-b_n)^2}{2h})} \; \\
	\end{split}
\end{equation}
In a supervised machine learning task, maximizing the correntropy between the model prediction and the true target exhibits exceptional robustness with respect to non-Gaussian noises, in particular to outliers, which refers to the \emph{maximum correntropy criterion} (MCC), because correntropy is a \emph{local} measure which is mainly determined by the Gaussian kernel function $k_h$ along $A=B$. It was also proved to extract more statistical moments from the data and has a close relation with the Renyi's entropy of the second order \cite{liu2007correntropy,principe2010information}.

\subsection{Noise Assumption Under MCC}
\label{subsec:mcc_noise}
We desire to rethink the noise assumption inherent in MCC. Utilizing MCC for the linear regression model with $N$ samples yields
\begin{equation}
	\begin{split}
		\label{equ:mcc_noise_1}
		\mathbf{w}&=arg\max _{\mathbf{w}} \frac{1}{N}\sum_{n=1}^{N}{\exp(-\frac{(t_n-\mathbf{x}_n\mathbf{w})^2}{2h})} \; \\
		&=arg\max _{\mathbf{w}} \frac{1}{N}\sum_{n=1}^{N}{\exp(-\frac{e_n^2}{2h})} \; \\
	\end{split}
\end{equation}
where $e_n\triangleq t_n-\mathbf{x}_n\mathbf{w}$ denotes the $n$-th prediction error. If we omit the fixed number $N$, we can find MCC will be equivalent to a multiplication form through an exponential function
\begin{equation}
	\begin{split}
		\label{equ:mcc_noise_2}
		\mathbf{w}&=arg\max _{\mathbf{w}}\sum_{n=1}^{N}{\exp(-\frac{e_n^2}{2h})} \; \\
		&=arg\max _{\mathbf{w}}\prod_{n=1}^{N}{\exp\{\exp(-\frac{e_n^2}{2h})\}} \; \\
		&=arg\max _{\mathbf{w}}\prod_{n=1}^{N}{\exp\{\exp(-\frac{(t_n-\mathbf{x}_n\mathbf{w})^2}{2h})\}} \; \\
	\end{split}
\end{equation}
which can be extraordinarily regarded as a likelihood function maximum if we assume independence for each $t_n$ and define the following PDF for the noise distribution
\begin{equation}
	\label{equ:mcc_likelihood}
	\mathcal{C}(e|0,h)\triangleq \exp\{\exp(-\frac{e^2}{2h})\}
\end{equation}
in which $\mathcal{C}(e|0,h)$ is defined as a correntropy-aware PDF over $e$ with a zero mean and the shape parameter $h$. Utilizing such an assumption on the noise distribution, we obtain the PDF of $t$ by $p(t|\mathbf{x})=\mathcal{C}(t|\mathbf{xw},h)$. Hence, assuming the independence for $t_n$, one can find the MLE based on the defined PDF $\mathcal{C}$ will be equivalent to the original MCC (\ref{equ:mcc_noise_2}).

\begin{figure}[t!]
	\centering
	\includegraphics[width=0.46\textwidth]{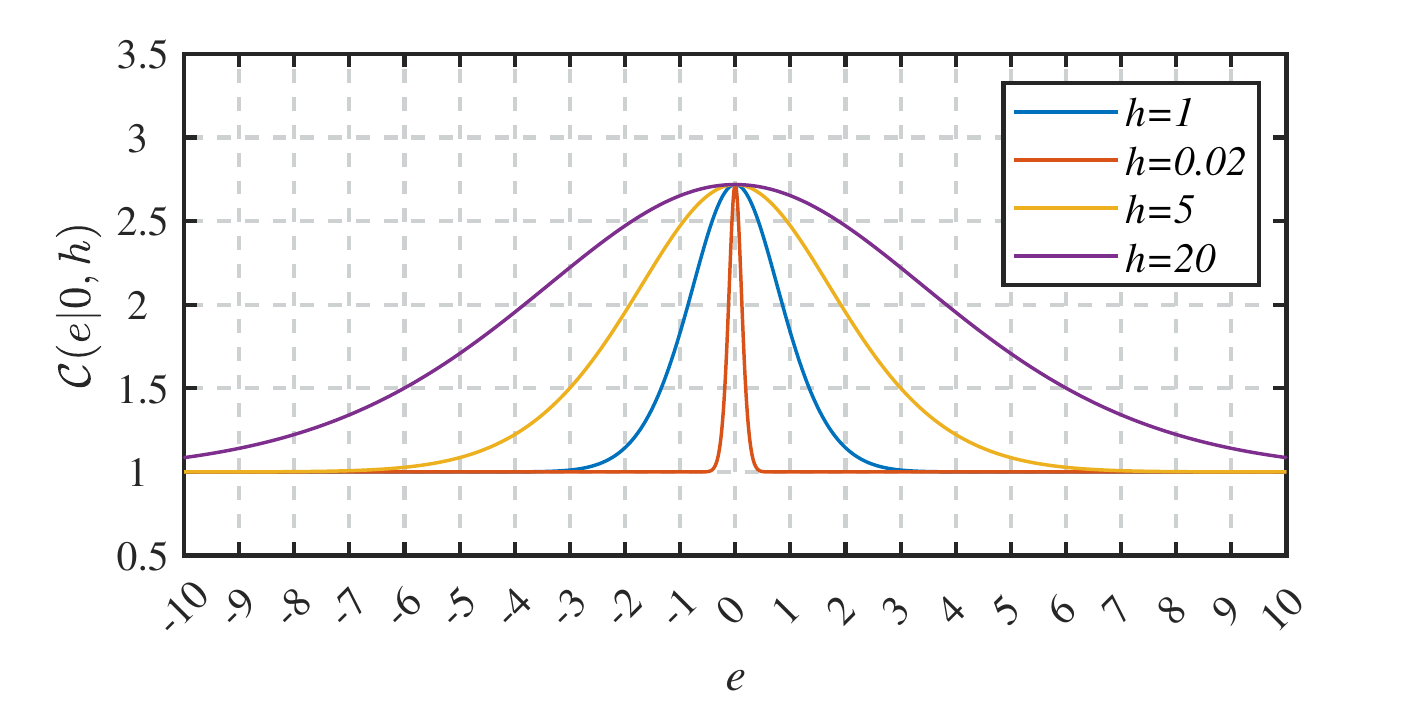}
	\caption{MCC-aware noise distribution $\mathcal{C}(e|0,h)$ with different $h$ values.}
	\label{fig_mcc_dist}
\end{figure}

It is important to investigate the property of the defined PDF $\mathcal{C}$. Unsurprisingly, it is not a `well-defined' PDF since one sees that its integral is infinite, thus, being an \emph{improper} distribution \cite{gelman1995bayesian}. Even more, when $e$ is far from the origin, the probability density defined by $\mathcal{C}(e|0,h)$ is close to $1$, rather than a normal case $0$, which seems to be a \emph{deviant} PDF. Nevertheless, in the present study, we demonstrate empirically that, such a \emph{deviant} MCC-aware noise distribution can largely improve the robust property for an ARD-based sparse regression model. We show some examples for $\mathcal{C}(e|0,h)$ in Fig. \ref{fig_mcc_dist} with different $h$ values. A further discussion for this \emph{deviant} noise assumption is given in Section \ref{subsec:likeli}.

\section{MCC-ARD for Robust Sparse Regression}
\label{sec:mcc-ard}

In this section, we desire to integrate the MCC-based robust regression with the ARD technique under a Bayesian inference framework, using the correntropy-aware noise assumption (\ref{equ:mcc_likelihood}) to derive the likelihood function, which is written by
\begin{equation}
	\begin{split}
		\label{equ:mcc_likelihood_function}
		p(\mathbf{t}|\mathbf{w},h)&=\prod_{n=1}^{N}\mathcal{C}(t_n|\mathbf{x}_n\mathbf{w},h) \; \\
		&=\prod_{n=1}^{N}\exp\{\exp(-\frac{(t_n-\mathbf{x}_n\mathbf{w})^2}{2h})\}   \; \\
	\end{split}
\end{equation}
However, one could find that the utilization of the MCC-aware likelihood function (\ref{equ:mcc_likelihood_function}) obstructs the analytical derivation for the posterior distribution $p(\mathbf{w}|\mathbf{t},\mathbf{a},h)$, contrast to $p(\mathbf{w}|\mathbf{t},\mathbf{a},\sigma^2)$ (\ref{equ:post_w}) under the Gaussian noise assumption, since the likelihood function (\ref{equ:mcc_likelihood_function}) is not conjugate with the Gaussian priors $p(\mathbf{w}|\mathbf{a})$ (\ref{equ:ard1}). Therefore, we resort to the variational Bayesian inference \cite{gelman1995bayesian}, which can approximate the posterior distribution for each variable. For simplicity, we can first treat the kernel bandwidth $h$ as a fixed parameter. Section \ref{subsec:kernel} gives a discussion about the treatment of $h$ as a random variable.

The variational Bayesian inference defines a surrogate PDF $Q(\mathbf{w},\mathbf{a})$ to approximate the posterior distribution $p(\mathbf{w},\mathbf{a}|\mathbf{t},h)$ which is furthermore assumed with the independence between $\mathbf{w}$ and $\mathbf{a}$ by $Q(\mathbf{w},\mathbf{a})=Q_{\mathbf{w}}(\mathbf{w})Q_{\mathbf{a}}(\mathbf{a})$, and tries to maximize the following free energy $F(Q_{\mathbf{w}}(\mathbf{w})Q_{\mathbf{a}}(\mathbf{a}))$
\begin{equation}
	\begin{split}
		\label{equ:mcc_free_energy}
		&F(Q_{\mathbf{w}}(\mathbf{w})Q_{\mathbf{a}}(\mathbf{a})) \triangleq \; \\
		&\int Q_{\mathbf{w}}(\mathbf{w})Q_{\mathbf{a}}(\mathbf{a})\log\frac{p(\mathbf{w},\mathbf{a},\mathbf{t},h)}{Q_{\mathbf{w}}(\mathbf{w})Q_{\mathbf{a}}(\mathbf{a})}d\mathbf{w}d\mathbf{a} \; \\
	\end{split}
\end{equation}
which is maximized when and only when $Q(\mathbf{w},\mathbf{a})$ is equal to the posterior distribution $p(\mathbf{w},\mathbf{a}|\mathbf{t},h)$. The logarithmic forms of $Q_{\mathbf{w}}(\mathbf{w})$ and $Q_{\mathbf{a}}(\mathbf{a})$ are expressed by
\begin{equation}
	\begin{split}
		\label{equ:qw_qa_1}
		\log Q_{\mathbf{w}}(\mathbf{w})&=\left<\log p(\mathbf{w},\mathbf{a},\mathbf{t},h)\right>_{Q_{\mathbf{a}}(\mathbf{a})} \; \\
		\log Q_{\mathbf{a}}(\mathbf{a})&=\left<\log p(\mathbf{w},\mathbf{a},\mathbf{t},h)\right>_{Q_{\mathbf{w}}(\mathbf{w})} \; \\
	\end{split}
\end{equation}
where $\left<\cdot\right>_Q$ means the expectation with respect to PDF $Q$. The log joint distribution $\log p(\mathbf{w},\mathbf{a},\mathbf{t},h)$ is
\begin{equation}
	\begin{split}
		\label{equ:log_joint}
		&\log p(\mathbf{w},\mathbf{a},\mathbf{t},h)=\log p(\mathbf{t}|\mathbf{w},h)+\log p(\mathbf{w}|\mathbf{a}) +\log p(\mathbf{a}) \; \\
		=&\sum_{n=1}^{N}\exp(-\frac{(t_n-\mathbf{x}_n\mathbf{w})^2}{2h})-\frac{1}{2}\mathbf{w}^T\mathbf{A}\mathbf{w}-\frac{1}{2}\log|\mathbf{A}|+const \; \\
	\end{split}
\end{equation}
Gathering the relevant terms with respect to $\mathbf{w}$ and $\mathbf{a}$, one then obtains
\begin{equation}
	\begin{split}
		\label{equ:qw_qa_2}
		\log Q_{\mathbf{w}}(\mathbf{w})&=\sum_{n=1}^{N}\exp(-\frac{(t_n-\mathbf{x}_n\mathbf{w})^2}{2h})-\frac{1}{2}\mathbf{w}^T\left<\mathbf{A}\right>_{Q_{\mathbf{a}}(\mathbf{a})}\mathbf{w} \; \\
		\log Q_{\mathbf{a}}(\mathbf{a})&=-\frac{1}{2}\sum_{d=1}^{D}a_d\left<w_d^2\right>_{Q_{\mathbf{w}}(\mathbf{w})}-\frac{1}{2}\sum_{d=1}^{D}\log a_d \; \\
	\end{split}
\end{equation}
However, one could see that $Q_{\mathbf{w}}(\mathbf{w})$ cannot be expressed with an analytical form. Therefore, we further utilize the Laplacian approximation to $\log Q_{\mathbf{w}}(\mathbf{w})$ through a quadratic form by
\begin{equation}
	\label{equ:qw_lap}
	\log Q_{\mathbf{w}}(\mathbf{w})\approx \log Q_{\mathbf{w}}(\mathbf{w}^*)-\frac{1}{2}(\mathbf{w}-\mathbf{w}^*)^T\mathbf{H}(\mathbf{w}^*)(\mathbf{w}-\mathbf{w}^*)
\end{equation}
in which $\mathbf{w}^*$ is the maximum point of $\log Q_{\mathbf{w}}(\mathbf{w})$, and $\mathbf{H}(\mathbf{w}^*)$ denotes the \emph{negative} Hessian matrix of $\log Q_{\mathbf{w}}(\mathbf{w})$ at $\mathbf{w}^*$
\begin{equation}
	\begin{split}
		\label{equ:negative_hessian}
		&\mathbf{H}(\mathbf{w})=-\frac{\partial^2\log Q_{\mathbf{w}}(\mathbf{w})}{\partial\mathbf{w}\partial\mathbf{w}^T} \; \\
		&=-\frac{1}{h}\sum_{n=1}^{N}{\mathbf{x}_n^T\left\{ \exp(-\frac{e_n^2}{2h})(\frac{e_n^2}{h}-1)\right\}\mathbf{x}_n}+\left<\mathbf{A}\right>_{Q_{\mathbf{a}}(\mathbf{a})} \; \\
	\end{split}
\end{equation}
Thus by approximating $\log Q_{\mathbf{w}}(\mathbf{w})$ with a quadratic form (\ref{equ:qw_lap}), $Q_{\mathbf{w}}(\mathbf{w})$ can be regarded as a Gaussian distribution $Q_{\mathbf{w}}(\mathbf{w})=\mathcal{N}(\mathbf{w}|\mathbf{w}^*,\mathbf{H}(\mathbf{w}^*)^{-1})$. The expectation $\left<w_d^2\right>$ can be calculated by
\begin{equation}
	\left<w_d^2\right>_{Q_{\mathbf{w}}(\mathbf{w})}=w_d^{*2}+s_d^2
\end{equation}
where $s_d^2$ is the $d$-th diagonal element in $\mathbf{H}(\mathbf{w}^*)^{-1}$. As a result, $\log Q_{\mathbf{a}}(\mathbf{a})$ can be expressed by
\begin{equation}
	\label{equ:log_qa}
	\log Q_{\mathbf{a}}(\mathbf{a})=-\frac{1}{2}\sum_{d=1}^{D}\{a_d(w_d^{*2}+s_d^2)+\log a_d\}
\end{equation}
through which $Q_{\mathbf{a}}(\mathbf{a})$ could be regarded to obey the following Gamma distribution
\begin{equation}
	\label{equ:a_gamma}
	Q_{\mathbf{a}}(\mathbf{a})=\prod_{d=1}^{D}{Q_{a_d}(a_d)}=\prod_{d=1}^{D}{\varGamma(a_d|a_d^*,\frac{1}{2})}
\end{equation}
where $\varGamma(a_d|a_d^*,\frac{1}{2})$ denotes a Gamma distribution over $a_d$ with the degree of freedom $\frac{1}{2}$ and the expectation $a_d^*$ that is
\begin{equation}
	\label{equ:slr_a_star}
	a_d^*=\frac{1}{w_d^{*2}+s_d^2}
\end{equation}
which can be in turn substituted into $\left<\mathbf{A}\right>_{Q_{\mathbf{a}}(\mathbf{a})}$ for $\log Q_{\mathbf{w}}(\mathbf{w})$ in (\ref{equ:qw_qa_2})-(\ref{equ:negative_hessian}).

By updating $\log Q_{\mathbf{w}}(\mathbf{w})$ and $\log Q_{\mathbf{a}}(\mathbf{a})$ alternately, the free energy $F(Q_{\mathbf{w}}(\mathbf{w})Q_{\mathbf{a}}(\mathbf{a}))$ will be maximized, so that one could obtain the MAP estimations for $\mathbf{w}$ and $\mathbf{a}$. To optimize $\mathbf{w}$, one can perceive that $\log Q_{\mathbf{w}}(\mathbf{w})$ is exactly equal to $L_2$-regularized MCC with the current $\mathbf{a}$ values (\ref{equ:qw_qa_2}), which could be effectively optimized by the fixed-point update with fast convergence \cite{chen2015convergence}
\begin{equation}
	\label{equ:w_fp}
	\mathbf{w}=(\mathbf{X}^T\mathbf{\Psi}\mathbf{X}+\mathbf{A})^{-1}\mathbf{X}^T\mathbf{\Psi}\mathbf{t}
\end{equation}
where $\mathbf{\Psi}$ is a $N\times N$ diagonal matrix with the diagonal element $\Psi_{nn}=\exp(-e_n^2/2h)$. By finding the maximum point $\mathbf{w}^*$ for $\log Q_{\mathbf{w}}(\mathbf{w})$, one could optimize $\mathbf{a}$ by (\ref{equ:slr_a_star}), while the following update could give faster convergence \cite{wipf2007new,tipping2001sparse}
\begin{equation}
	\label{equ:a_fast}
	a_d^*=\frac{1-a_d^*s_d^2}{w_d^{*2}}
\end{equation}
which could be regarded as a fixed-point form of (\ref{equ:slr_a_star}). During the training, some $a_d$ will diverge to infinity, as introduced in Section \ref{sec:lsr_ard}. We could employ an upper threshold and prune the corresponding features if their relevance parameter $a_d$ exceeds this upper limit. The proposed MCC-ARD method for robust sparse regression is summarized in Algorithm \ref{cslr}.

\begin{algorithm}[h]
	\caption{MCC-ARD Algorithm}
	\label{cslr}
	\begin{algorithmic}[1]
		\State \textbf{input}:
		
		training samples $\{\mathbf{x}_n,t_n\}_{n=1}^N$;
		
		kernel bandwidth $h$;
		
		threshold for relevance parameter $a_{\max}$;
		
		\State \textbf{initialize}:
		
		model parameter $w_d$ ($d=1,\cdots,D$);
		
		relevance parameter $a_d=1$ ($d=1,\cdots,D$);
		
		\State \textbf{output}:
		
		model parameter $w_d$ ($d=1,\cdots,D$)
		\Repeat 
		\State $\mathbf{w}$-step: update $\mathbf{w}$ according to (\ref{equ:w_fp});
		\State $\mathbf{a}$-step: update $\mathbf{a}$ according to (\ref{equ:a_fast});
		\If{$a_d\geqslant a_{\max}$}
		\State set the corresponding $w_d$ to zero and prune this dimension in the following updates
		\EndIf
		\Until the number of iterations is larger than an upper limit or the parameter change is small enough
	\end{algorithmic}
\end{algorithm}

\section{Experiments}
\label{sec:exp}
We assess the proposed MCC-ARD algorithm by a synthetic dataset, comparing it with the conventional ARD-based sparse regression introduced in Section \ref{sec:lsr_ard} (denoted by LS-ARD), and the $L_1$-regularized MCC \cite{he2010maximum,he2011regularized,he2013half} (MCC-$L_1$) optimized with an EM method \cite{figueiredo2003adaptive,schmidt2007fast}. The kernel bandwidth $h$ for both MCC- ARD and MCC-$L_1$ are selected by cross validation, while the latter uses another cross validation for regularization parameter $\lambda$. The pruning threshold $a_{\max}$ is set as $10^6$ for both LS-ARD and MCC-ARD.

\begin{figure}[t!]
	\centering
	\includegraphics[width=0.48\textwidth]{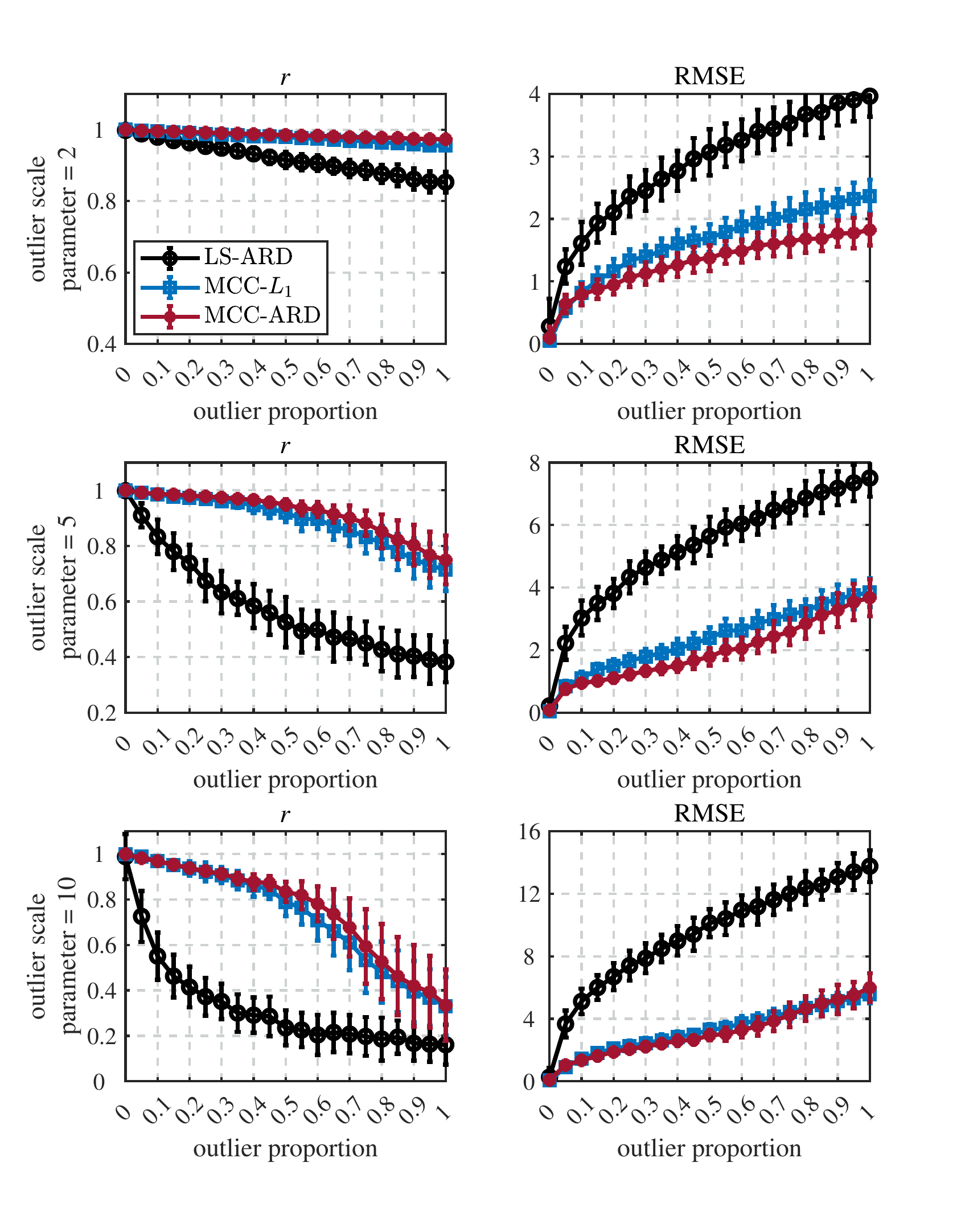}
	\caption{Correlation coefficient ($r$) and root mean squared error (RMSE) with the noisy and high-dimensional dataset under different outlier proportions and scale parameters. The results are averaged across 100 Monte-Carlo repetitions where the error bars represent the standard deviations.}
	\label{fig_toy}
\end{figure}

We generate a noisy and high-dimensional synthetic dataset with the following method. We first generate 300 i.i.d. training samples and 300 i.i.d. testing samples, which obey the $1000$- dimensional standard normal distribution. To obtain the model output, we employ a sparse true solution $\mathbf{w}^*$ which is a $1000$-dimensional vector where only the first $30$ dimensions are non- zero components and the other $970$ components are zero
\begin{equation}
	\label{equ:true_solution}
	\mathbf{w}^*=[\overset{1000\; dimensions}{\overbrace{w^*_1,w^*_2,\cdots,w^*_{30},\underset{970\; zero-components}{\underbrace{0,0,0,0,0,0,\cdots,0}}}}]^T
\end{equation}
in which the non-zero elements were randomly generated from the univariate standard normal distribution. The model output is obtained with the linear regression model (\ref{equ:lr}). To assess the robustness of each algorithm, we use the following distribution for the additive noise term $\epsilon$
\begin{equation}
	\label{equ:noise}
	\epsilon \sim (1-\psi)\mathcal{N}(\epsilon|0,0.05)+\psi\mathcal{L}(\epsilon|0,\tau)
\end{equation}
in which $\mathcal{L}(\epsilon|0,\tau)$ denotes the Laplace distribution over $\epsilon$ with zero mean and the scale parameter $\tau$ to imitate outliers, and $\psi$ means the proportion of outliers among the additive noise. We employ a popular setting for robustness evaluation, where only the training dataset is contaminated with the above corruption, whereas the noise term is excluded for the testing data, as was advised in \cite{zhu2004class}. We consider the following values for the scale parameter $\tau$: $2$, $5$, and $10$, indicating increasing strengths for the outliers. The outlier proportion $\psi$ is increased from 0 to 1.0 with a step 0.05. The regression performance is evaluated by two classical regression performance indicators, correlation coefficient ($r$) and root mean squared error (RMSE), which are computed respectively by
\begin{equation}
	\begin{split}
		\label{equ:pi}
		r=Cov(\mathbf{\hat{t}},\mathbf{t})/&\sqrt{Var(\mathbf{\hat{t}})Var(\mathbf{t})}  \; \\
		\text{RMSE}=&\sqrt{\frac{1}{N}\lVert \mathbf{\hat{t}} - \mathbf{t} \rVert^2}  \; \\
	\end{split}
\end{equation}
where $Cov(\cdot,\cdot)$ and $Var(\cdot)$ mean the covariance and variance, respectively, while $\mathbf{\hat{t}}$ is the collection of the model predictions. We present the prediction performance of each algorithm with the above simulation settings with 100 Monte-Carlo repetitions in Fig. \ref{fig_toy}, which exhibit a noisy and high-dimensional dataset. One could observe that, the proposed MCC-ARD outperforms the conventional LS-ARD largely by significantly higher $r$ and lower RMSE, when the high-dimensional data is contaminated by the non-Gaussian noises under each scale parameter $\tau$. One further perceives that the proposed MCC-ARD achieves higher $r$ than the existing MCC-$L_1$ under each scale parameter $\tau$, and lower RMSE for $\tau=2$ and $5$. MCC-ARD and MCC-$L_1$ give similar RMSE when $\tau=10$. When $\tau$ becomes larger than $10$, the conclusion of performance comparison is analogous to the case when $\tau$ is equal to $10$. We would like to remind here that the proposed MCC-ARD method only has one hyperparameter $h$ to be tuned carefully, whereas MCC-$L_1$ needs to adjust two important hyper-parameters, namely, the kernel size $h$ and the regularization parameter $\lambda$.

\begin{figure}[t!]
	\centering
	\includegraphics[width=0.36\textwidth]{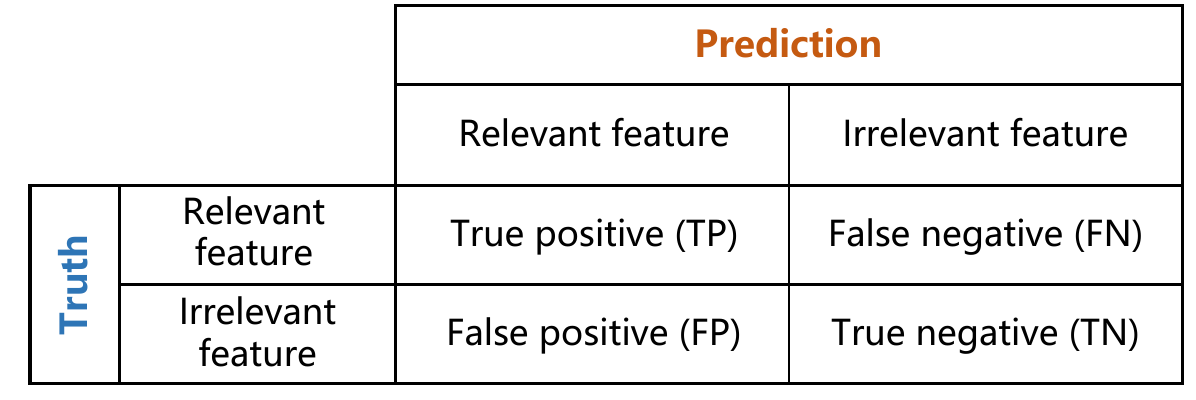}
	\caption{Confusion matrix for the feature selection problem.}
	\label{fig_con_mat}
\end{figure}

\begin{figure}[t!]
	\centering
	\includegraphics[width=0.48\textwidth]{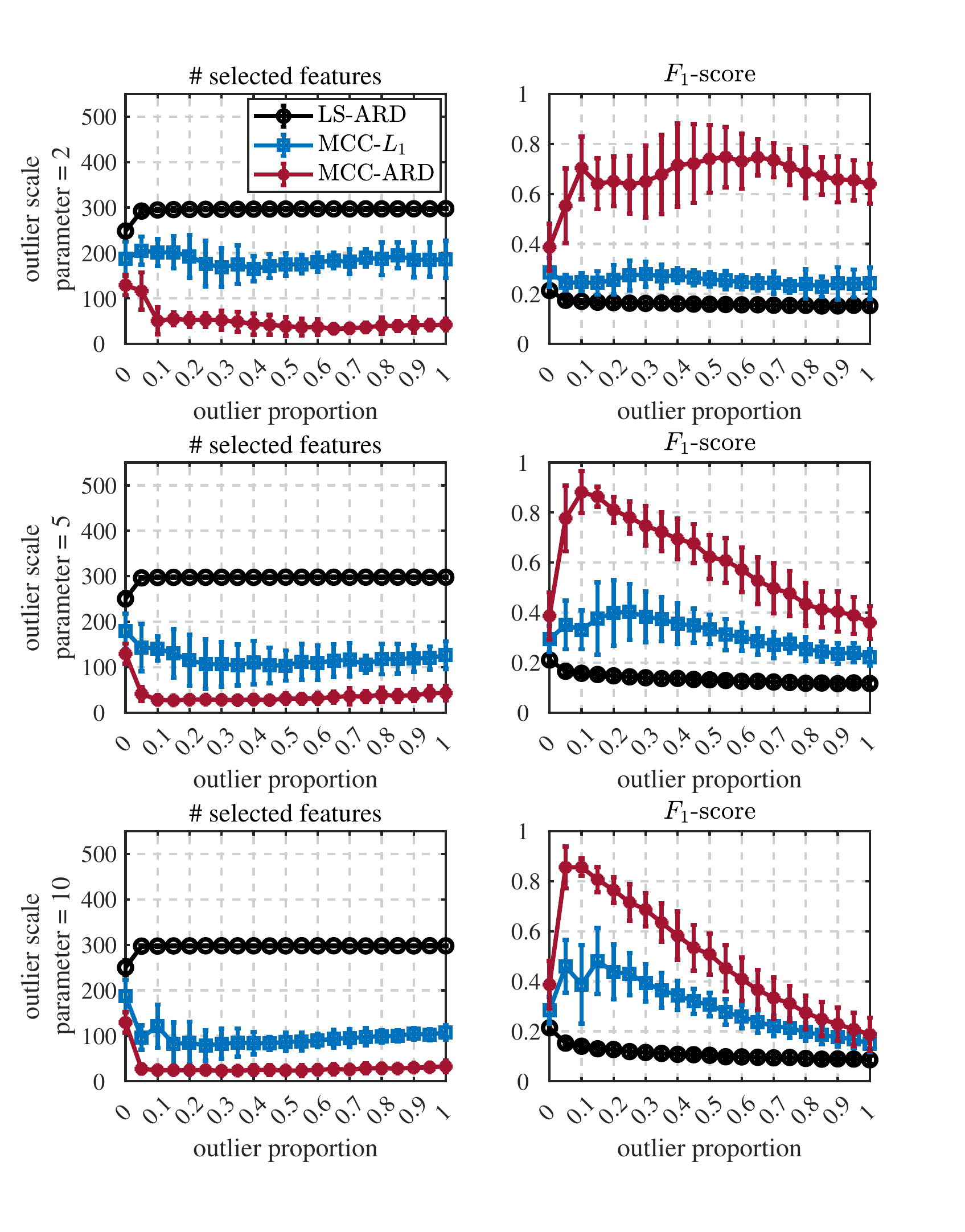}
	\caption{Number of selected features and F1-score of each regression algorithm for the high-dimensional dataset.}
	\label{fig_toy_fea}
\end{figure}

On the other hand, we also consider the feature selection of the high-dimensional dataset in the presence of outliers, where we can evaluate the selection quality quantitatively because the ground-truth `relevant'/`irrelevant' label for each dimension is known. The feature selection can be viewed as an unbalanced classification task in which we have 30 `relevant' features and 970 `irrelevant' features. In the trained regression models, the pruned dimensions are predicted as `irrelevant' features, while the retained ones with non-zero model parameters are regarded as `relevant'. The confusion matrix for this classification issue is illustrated in Fig. \ref{fig_con_mat}. We utilize a comprehensive performance indicator, F1-score, to evaluate this unbalanced problem
\begin{equation}
	\begin{split}
		F_1&=2\times\frac{Precision\times Recall}{Precision+Recall} \;\\
	\end{split}
\end{equation}
which is the harmonic mean of $Precision=TP/(TP+FP)$ and $Recall=TP/(TP+FN)$. Fig. \ref{fig_toy_fea} illustrates the number of selected features and F1-score of feature selection for each algorithm. One can observe that when the data is contaminated by the outliers, the number of selected features by MCC-ARD is closer to the ground truth of $30$ relevant features, compared with the conventional LS-ARD and existing MCC-$L_1$. Notably MCC-ARD reveals significantly higher F1-score in the feature selection than other two algorithms in the presence of outliers, showing exceptional feature selection capability in a noisy and high-dimensional scenario. Even more, MCC-ARD also gives higher F1-score without outlier contamination (proportion=$0$). Remarkably, when the outlier scale parameter equals $5$ or $10$, a small outlier proportion (e.g. $0.05$) improves largely the F1- score for feature selection for the proposed MCC-ARD, which seems rather surprising and necessitates a further investigation to interpret this effect.

\section{Discussion}
\label{sec:disc}

\subsection{MCC-Aware Noise Assumption}
\label{subsec:likeli}
\begin{figure}[t!]
	\centering
	\includegraphics[width=0.46\textwidth]{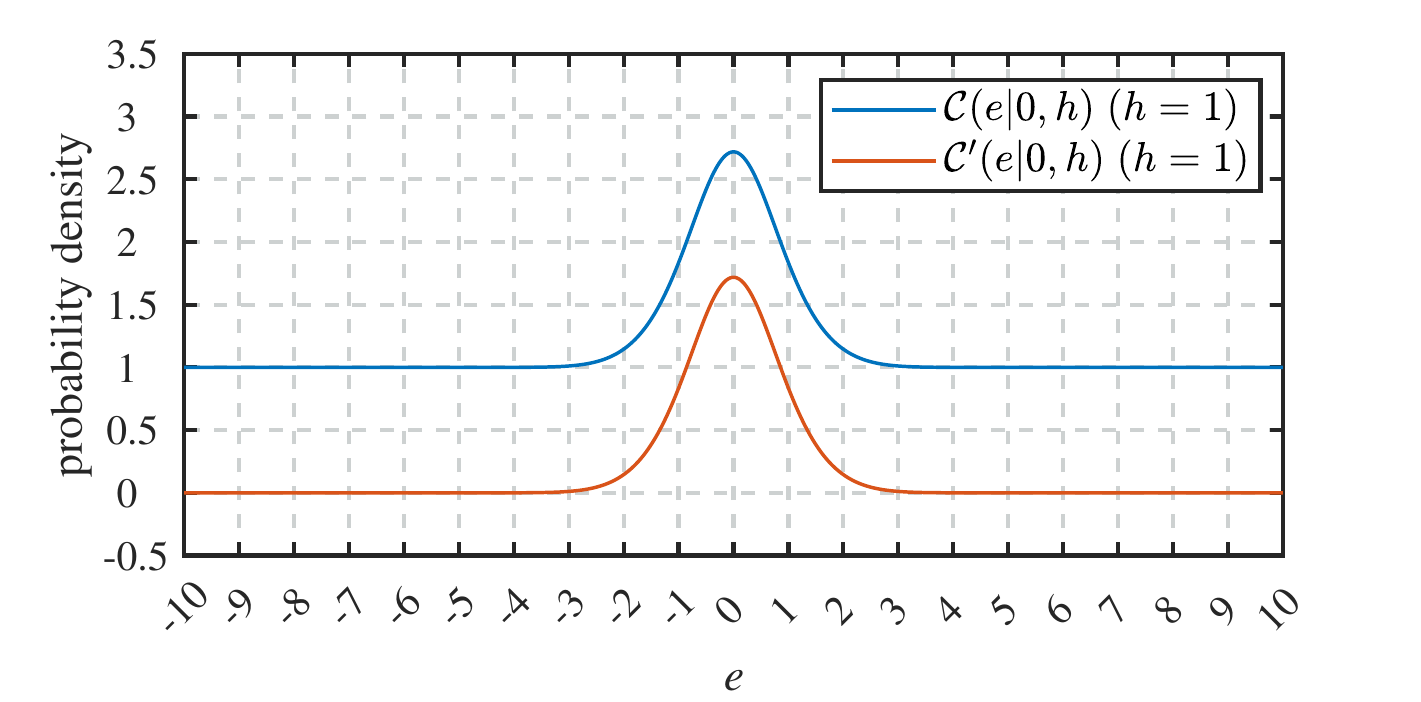}
	\caption{Comparison between \emph{deviant} $\mathcal{C}(e|0,h)$ and \emph{proper} $\mathcal{C}'(e|0,h)$.}
	\label{fig_mcc_dist_2}
\end{figure}

\begin{figure}[t!]
	\centering
	\includegraphics[width=0.48\textwidth]{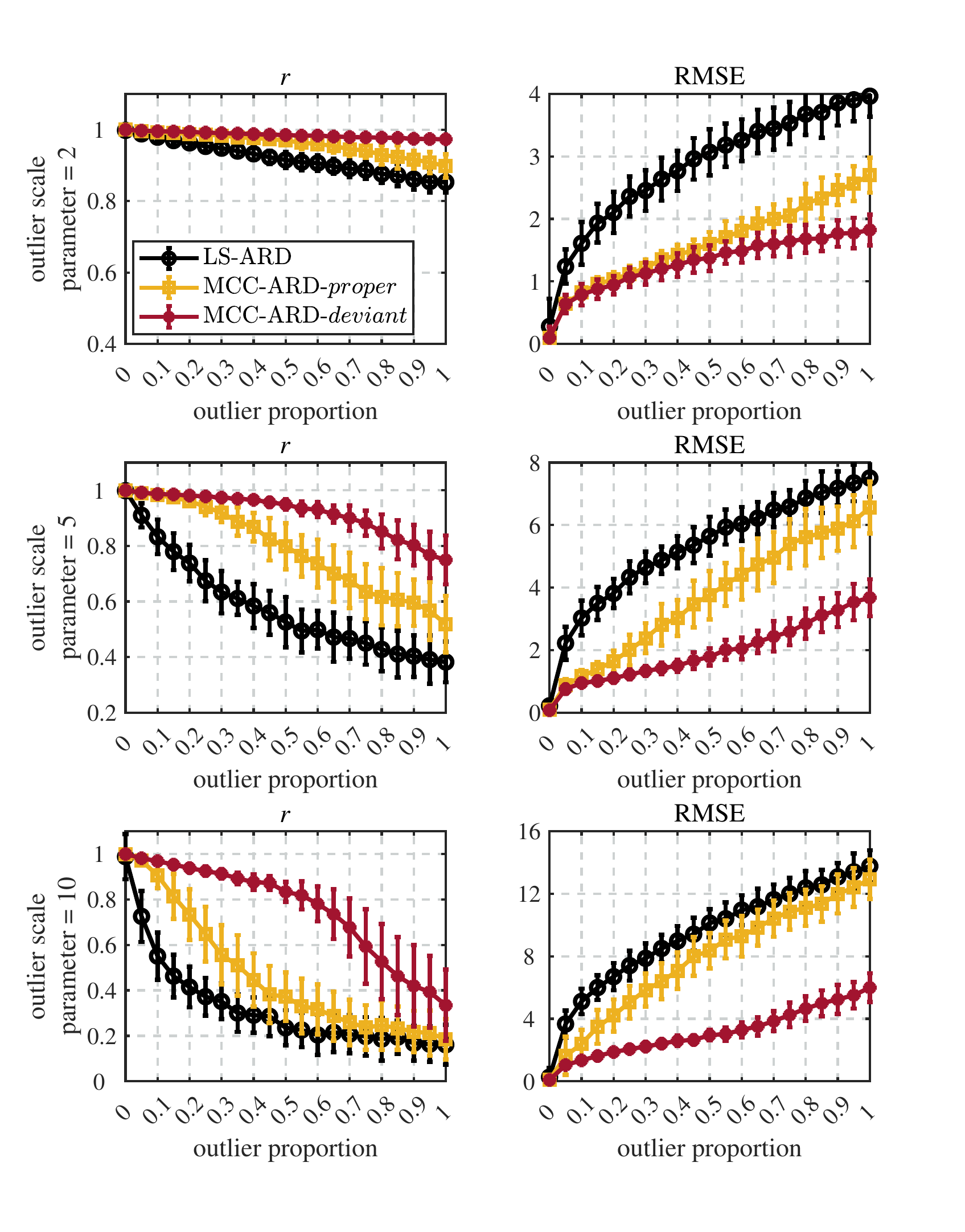}
	\caption{Correlation coefficient ($r$) and root mean squared error (RMSE) for the MCC-ARD regression algorithms which are derived by the \emph{proper} $\mathcal{C}'(e|0,h)$ and the \emph{deviant} $\mathcal{C}(e|0,h)$, respectively.}
	\label{fig_toy_2}
\end{figure}

It is indispensable to discuss whether the MCC-aware noise assumption $\mathcal{C}(e|0,h)$ (\ref{equ:mcc_likelihood}) is adequate to be utilized in a robust regression model from a Bayesian perspective. Conventionally, an \emph{improper} distribution, referring to a non-normalizable PDF, can be only permitted for a prior distribution (and the resultant posterior distribution) in a classical Bayesian regime \cite{gelman1995bayesian}. The likelihood function (equally the noise distribution), to the best of our knowledge, has for the first time been utilized with such a \emph{deviant} distribution $\mathcal{C}(e|0,h)$, which does not even converge to $0$ far from the origin. To verify the validity of such a \emph{deviant} noise assumption, we define the following noise distribution
\begin{equation}
	\label{equ:mcc_likelihood_2}
	\mathcal{C}'(e|0,h)\triangleq \exp\{\exp(-\frac{e^2}{2h})\}-1
\end{equation}
which is a simple translation of $\mathcal{C}(e|0,h)$ towards the horizontal axis, and can be proved a normalizable PDF by elementary derivation, shown in Fig. \ref{fig_mcc_dist_2}. With this \emph{proper} noise distribution, we conduct a similar derivation as in Section \ref{sec:mcc-ard}, and compare the experimental results utilizing the identical synthetic dataset from Section \ref{sec:exp} in Fig. \ref{fig_toy_2}. One can observe that, for each outlier scale parameter, the \emph{deviant} MCC-ARD outperforms evidently the \emph{proper} one. In particular, when the outlier scale parameter is $10$, the \emph{proper} MCC-ARD even achieves similar results with the conventional LS-ARD, showing poor robustness compared with the \emph{deviant} one. Therefore, the validity of the MCC-aware \emph{deviant} noise distribution $\mathcal{C}(e|0,h)$ (\ref{equ:mcc_likelihood}) is empirically proved. The robustness of $\mathcal{C}(e|0,h)$, in our opinion, can be interpreted heuristically as follows.

The prominent characteristic of the \emph{deviant} $\mathcal{C}(e|0,h)$ is that, its probability density reaches the maximum at the origin while it converges to $1$ when $e\rightarrow\infty$. In the usual noise assumptions (e.g. Gaussian), the probability density converges to $0$ when $e$ is arbitrarily large, which seems to be a reasonable hypothesis. However, if a dataset is in particular prone to adverse outliers, this hypothesis would be unreliable, because some errors with large values do happen, indicating non-zero probability density even though far from the origin. By comparison, our \emph{deviant} $\mathcal{C}(e|0,h)$ precisely assumes non-zero density for the arbitrarily large error. Thus, we would like to argue that the MCC-aware $\mathcal{C}(e|0,h)$ is a more rational noise assumption when the dataset is prone to outliers, as was demonstrated by the experimental results. To the best of our knowledge, this is the first time that the exceptional robustness of MCC has been interpreted from the perspective of noise assumption. Further investigations are being studied for more solid theoretical guarantees.

\subsection{Kernel Bandwidth Determination}
\label{subsec:kernel}
In this paper, we determined the kernel bandwidth $h$ through cross validation, which is a widely employed strategy for MCC based algorithms \cite{liu2007correntropy,feng2015learning,chen2016generalized,ma2018bias}. Although the kernel bandwidth $h$ could be computed directly from the kernel density estimation, such as \emph{Silverman's Rule} \cite{silverman2018density}, it was reported to result in poor consequence in \cite{he2011regularized}. We desire to investigate how to treat this hyper-parameter as a random variable and integrate it with the Bayesian inference as well. Using the non-informative hyper- prior for $h$ yields the log joint distribution $\log p(\mathbf{w},\mathbf{a},\mathbf{t},h)$
\begin{equation}
	\begin{split}
		\label{equ:log_joint_new}
		&\log p(\mathbf{w},\mathbf{a},\mathbf{t},h) \; \\
		=&\log p(\mathbf{t}|\mathbf{w},h)+\log p(\mathbf{w}|\mathbf{a}) +\log p(\mathbf{a})+\log p(h) \; \\
		=&\sum_{n=1}^{N}\exp(-\frac{(t_n-\mathbf{x}_n\mathbf{w})^2}{2h})-\frac{1}{2}\mathbf{w}^T\mathbf{A}\mathbf{w}-\frac{1}{2}\log|\mathbf{A}|-\log h \; \\
	\end{split}
\end{equation}
Accordingly, the variational inference becomes
\begin{equation}
	\begin{split}
		\label{equ:qw_new}
		&\log Q_{\mathbf{w}}(\mathbf{w})=\left<\log p(\mathbf{w},\mathbf{a},\mathbf{t},h)\right>_{Q_{\mathbf{a}}(\mathbf{a})Q_{h}(h)} \; \\
		&=\sum_{n=1}^{N}\left<\exp(-\frac{e_n^2}{2h})\right>_{Q_{h}(h)}-\frac{1}{2}\mathbf{w}^T\left<\mathbf{A}\right>_{Q_{\mathbf{a}}(\mathbf{a})}\mathbf{w} \; \\
	\end{split}
\end{equation}
\begin{equation}
	\begin{split}
		\label{equ:qa_new}
		&\log Q_{\mathbf{a}}(\mathbf{a})=\left<\log p(\mathbf{w},\mathbf{a},\mathbf{t},h)\right>_{Q_{\mathbf{w}}(\mathbf{w})Q_{h}(h)} \; \\
		&=-\frac{1}{2}\sum_{d=1}^{D}a_d\left<w_d^2\right>_{Q_{\mathbf{w}}(\mathbf{w})}-\frac{1}{2}\sum_{d=1}^{D}\log a_d \; \\
	\end{split}
\end{equation}
\begin{equation}
	\begin{split}
		\label{equ:qh_new}
		&\log Q_{h}(h)=\left<\log p(\mathbf{w},\mathbf{a},\mathbf{t},h)\right>_{Q_{\mathbf{w}}(\mathbf{w})Q_{\mathbf{a}}(\mathbf{a})} \; \\
		&=\sum_{n=1}^{N}\left<\exp(-\frac{(t_n-\mathbf{x}_n\mathbf{w})^2}{2h})\right>_{Q_{\mathbf{w}}(\mathbf{w})}-\log h \; \\
	\end{split}
\end{equation}
where, however, one can find that the expectations with respect to the correntropy term in $\log Q_{\mathbf{w}}(\mathbf{w})$ and $\log Q_{h}(h)$ is pretty hard to compute analytically. Thus, some other approximations are essential to treat the bandwidth $h$ as a random variable. In our future work, we will do a deeper exploration so that MCC will be implemented with `\emph{adaptive robustness}' and `\emph{adaptive sparseness}', integrated with the ARD technique in a Bayesian framework.

\section{Conclusion}
\label{sec:con}
In this paper, we expose the inherent noise assumption under the MCC-based regression, and derive an explicit MCC-aware likelihood function. Integrated with the ARD technique, MCC- based robust regression can be implemented with the `\emph{adaptive sparseness}', where one does not need to tune the regularization hyperparameter. Compared with the conventional LS-ARD and the existing MCC-$L_1$, the proposed MCC-ARD algorithm can realize superior regression and feature selection in a noisy and high-dimensional scenario. Further investigations, including a Bayesian treatment of kernel bandwidth $h$ and an interpretation about the \emph{deviant} noise assumption $\mathcal{C}(e|0,h)$, will be explored in our future works.

\bibliography{bibli}
\bibliographystyle{IEEEtran}
\end{document}